\documentclass[review]{elsarticle}

\usepackage{lineno,hyperref}
\usepackage{epsfig}
\usepackage{graphicx}
\usepackage{multirow}
\usepackage{amsmath,amssymb}
\usepackage{color}
\modulolinenumbers[5]
\usepackage[numbers]{natbib}
\usepackage{algorithm}
\usepackage[noend]{algpseudocode}
\usepackage{soul,color}
\soulregister\cite7
\soulregister\ref7
\soulregister\pageref7
\usepackage{amsfonts}
 
\usepackage{epsfig}
\usepackage{graphicx}
\usepackage{amsmath}
\usepackage{amssymb}
\usepackage{mathtools}
\usepackage{resizegather}
\usepackage{marvosym}
\usepackage{dsfont}
\usepackage{enumitem}
\usepackage{caption}
\usepackage{multirow}
\usepackage{amsthm}
\theoremstyle{plain}

\newtheorem{proposition}{Proposition}

\usepackage{caption}

\usepackage{amsmath}
\DeclareMathOperator*{\argmax}{arg\,max}

\journal{International Journal of Intelligent Systems}

\begin{document}

\begin{frontmatter}

\title{Constraining Pseudo-label in Self-training Unsupervised Domain Adaptation with Energy-based Model}

\author[my2address]{Lingsheng Kong}

\author[my1address,my3address]{Bo Hu}

\author[my1address,my4address]{Xiongchang Liu}

\author[my1address,my5address]{Jun Lu}

\author[my6address]{Jane You}

\author[my1address]{Xiaofeng Liu*}
\ead{xliu61@mgh.harvard.edu; liuxiaofeng@cmu.edu}

\address[my1address]{Harvard University, Cambridge, MA, 02138 USA}
\address[my2address]{CIOMP, Chinese Academy of Sciences, CAS, China.}

\address[my3address]{National University of Singapore, Singapore.}

\address[my4address]{Dept. of Info. and EE, China University of Mining and Technology, China.}

\address[my5address]{Beth Israel Deaconess Medical Center and Harvard Medical School, Boston, MA, USA}

\address[my6address]{Dept. of Computing, The Hong Kong Polytechnic University, Hong Kong}

\begin{abstract}
Deep learning is usually data starved, and the unsupervised domain adaptation (UDA) is developed to introduce the knowledge in the labeled source domain to the unlabeled target domain. Recently, deep self-training presents a powerful means for UDA, involving an iterative process of predicting the target domain and then taking the confident predictions as hard pseudo-labels for retraining. However, the pseudo-labels are usually unreliable, thus easily leading to deviated solutions with propagated errors. In this paper, we resort to the energy-based model and constrain the training of the unlabeled target sample with an energy function minimization objective. It can be achieved via a simple additional regularization or an energy-based loss. This framework allows us to gain the benefits of the energy-based model, while retaining strong discriminative performance following a plug-and-play fashion. The convergence property and its connection with classification expectation minimization are investigated. We deliver extensive experiments on the most popular and large-scale UDA benchmarks of image classification as well as semantic segmentation to demonstrate its generality and effectiveness.
\end{abstract}


\end{frontmatter}


~\\~

\section{Introduction}

Deep learning (DL) has demonstrated its effectiveness in many areas, while it is usually data-starved and relies on the $i.i.d$ assumption of training and testing data. However, there are usually exists a discrepancy when we deploy the model in a new environment. The collection of a large number of target domain data for deep learning is challenging for many applications (e.g., medical image analysis), thereby hindering the wide adoption of deep learning methods \cite{wang2019domain,li2020discriminative}. For example, densely annotating a Cityscapes image on average takes about 90 minutes \cite{cordts2016cityscapes}, severely hindering the generalization of an autonomous driving system in different cities. Therefore, the unsupervised domain adaptation (UDA) is proposed to transfer knowledge from one labeled source domain to another target domain with the aid of unlabeled target data \cite{yang2018learning,yang2021learning,liu2021subtype}.

To this end, several solutions have been explored, e.g., feature/image-level adversarial training and maximum mean discrepancy \cite{li2020simplified,jia2020transferable}. Recently, one of the promising methods in UDA is the self-training \cite{lee2013pseudo,zou2019confidence}, which iteratively generates a set of one-hot pseudo-labels in the target domain, and then retrains the network according to these pseudo-labels of the unlabeled target data. The typical choice of the recent deep self-training usually be the one-hot pseudo-label (i.e., hard label), and demonstrating that it is essentially an entropy minimization process \cite{lee2013pseudo}, which enforces the output of the network to be as sharp as the one-hot hard pseudo-label.

\begin{figure}[t]
\centering
\includegraphics[width=8.5cm]{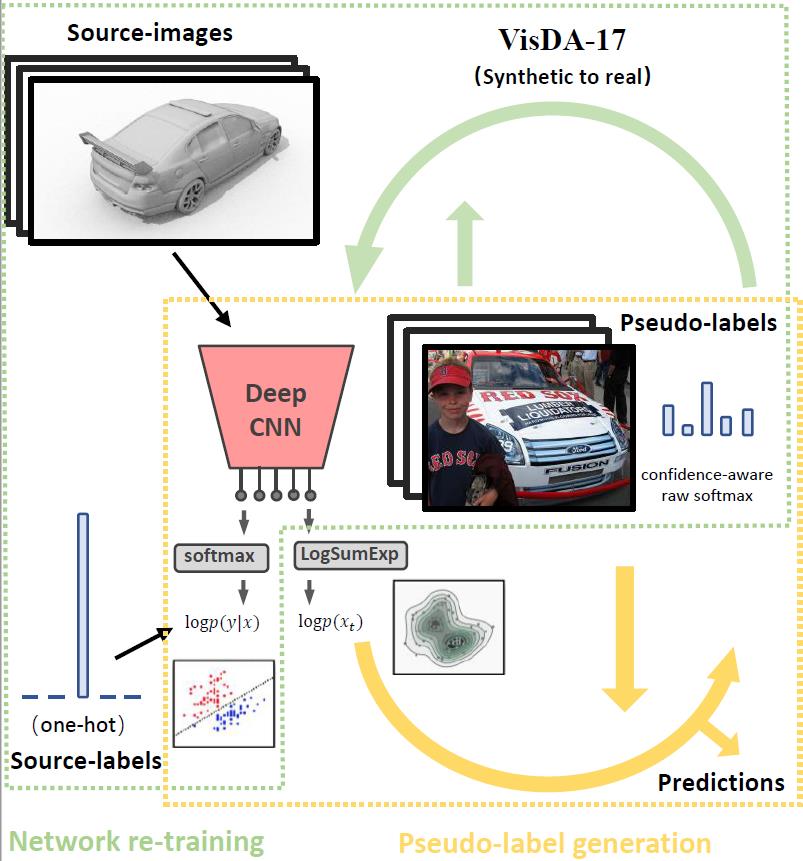}\\
\caption{The illustration of our Energy-constrained Self-training framework for UDA. Minimizing the pseudo label-irrelevant energy of $E_{\mathbf{w}}(\mathbf{x}_t)$ is introduced as additional objective for the target sample.}\label{fig:1} 
\end{figure}

However, the correctness of the generated pseudo-label is largely uncontrolled and can be hard to guarantee. Actually, the state-of-the-art accuracy of many UDA benchmarks is approximately 50\%. Especially in the first few epochs, it is hard to produce a reliable pseudo-label. In addition, the labels of natural images can also be highly ambiguous. Taking a sample image from VisDA17 \cite{peng2018visda} (see Fig. \ref{fig:1}) for illustration. We can see that both the person and car dominate significant portions in this image. Enforcing a model to be very confident in only one of the classes during training can also hurt the learning behavior \cite{bagherinezhad2018label}, particularly within the context of no ground truth label for the target samples. Encoding the pseudo-label as the one hot hard label vector is essentially trusting all of the pseudo-label as the ``ground truth”, which can lead to overconfident mistakes and propagated errors \cite{zou2019confidence}.

Targeting for this issue, our previous work \cite{zou2019confidence} proposes to make the network more conservative w.r.t. the pseudo-label of target sample via the manually defined pseudo label smoothing or entropy minimization. Specifically, \cite{zou2019confidence} proposes to modify the possibly inaccurate pseudo-label histogram distribution to be more smooth following a fixed smoothing operation for any data sample. For instance, the conventional self-training only transform the three-classes prediction $[0.3,0.6,0.2]$ to the one-hot pseudo-label $[0,1,0]$ with an argmax operation. Therefore, we can further apply the label smoothing to revise $[0,1,0]$ to $[0.05,0.9,0.05]$ for the subsequent conservative self-training. With the conventional cross-entropy loss, the confidence of the label can be reduced, which aligns with the practical reliability of pseudo-label. However, these constraints for pseudo-label can not be adaptively adjusted for different inputs or network parameters. It is clear that the reliability of the pseudo-label of the sample in Fig. \ref{fig:1} can be significantly lower than the normal samples. Besides, the pseudo-label produced in the later stage can be more trustable than the early stage if the training goes smoothly. Moreover, the label smoothing is explicitly dependent on the inaccurate pseudo-label, since it only revises the weight of histogram distribution. Actually, if the model with the current parameters is sufficiently confident to a specific sample, it is more reasonable to trust its pseudo label.

The difficulties mentioned above motivate us to develop an adaptive regularizer rather than the pre-defined pseudo label smoothing rules. We would expect our regularization signal for the target sample can adaptively reflect the current input and the network parameters in this training iteration rather than depend on the inaccurate pseudo label. 

{Recently, the correlation between the standard discriminative classifier and the energy-based model (EBM) is revealed \cite{grathwohl2019your}. Essentially, the standard discriminative classifier can be reinterpreted as an EBM for the joint distribution $p(\mathbf{x},\mathbf{y})$. Both the energy-based model and supervised discriminative model can be benefited from simultaneously optimizing both $p(\mathbf{y}|\mathbf{x})$ with cross-entropy (CE) loss and optimizing ${\rm log}p(\mathbf{x})$ with EBM \cite{grathwohl2019your}, respectively. Here, we demonstrate that optimizing ${\rm log}p(\mathbf{x})$ can be even more promising on self-training-based UDA. Since we do not have reliable labels for C.E. loss training on the target domain, and ${\rm log}p(\mathbf{x})$ can be an idea regularization signal. We note that ${\rm log}p(\mathbf{x})$ is explicitly correlated with the input $\mathbf{x}$ and network parameter w.r.t. $p(\cdot)$, and independent to pseudo label $\mathbf{y}$.}

Therefore, we propose a simple and straightforward idea that configures the energy minimization of data point $\mathbf{x}$ (i.e., ${\rm log}p(\mathbf{x})$) as an additional regularization term. The target domain examples are optimized with pseudo-label C.E. loss and EBM objective. With the help of the energy-based model, our self-training is expected to be more controllable. We give a thorough convergence analysis and theoretically connect it with classification expectation-maximization (CEM).


{In addition to the EBM-based regularization, we further propose an efficient EBM loss for the target sample, which is able to integrate the confidence-aware pseudo-label into the energy minimization objective. Based on the insights that the EBMs capture dependencies by associating a scalar energy value (i.e., a measure of compatibility) to each configuration of the variables (e.g., $\mathbf{y}$ and $\mathbf{x}$), we should not minimize the energy of all possible $\mathbf{y}$ and $\mathbf{x}$ configuration. It is more reasonable to focus on minimizing the correct energy w.r.t. configuration of $\mathbf{y}$ and $\mathbf{x}$. In the context of self-training UDA, the unprocessed soft pseudo-label can be a good signal as the weights of energy at each class. The histogram distribution value can also inherit the confidence of the discriminator, which provides more information for the later round training. Besides, the pseudo-label of Fig. \ref{fig:1} can be a multi-modal distribution which is also able to alleviate the adverse effect of the confusing images. It not only speeds up the training, removes the hyper-parameter validation but also achieves a comparable or even better performance than EBM regularization.}

This work extends our previous paper \cite{Liu2021energy} in the following significant ways:

\textbf{$\bullet$} {We propose a practical and scalable EBM loss for the target sample with the confidence-aware soft pseudo-label. It can efficiently speed up the training and remove the hyper-parameter validation, while not sacrificing the performance.}

\textbf{$\bullet$} {We theoretically analyze the convergence property of our energy-constrained self-training. Besides, its connection with the classification expectation maximization (CEM) is investigated from the probabilistic perspective.}

In addition, we further demonstrate the generality of our methods (both EBM regularizer and EBM loss) in more UDA benchmarks (e.g., Office-31 and SYNTHIA2Cityscapes) with more visualization and analysis.

The main contributions of this paper are summarized as

\textbf{$\bullet$} We propose a novel and intuitive framework to incorporate the EBM into the self-training UDA as a pseudo-label-irrelevant regularization signal, which benefits from both the generative and the discriminative model.   

\textbf{$\bullet$} It can simply be formulated as a regularization term of hard pseudo-label supervised training, and can also be an EBM loss for the target sample with the confidence-aware soft pseudo-label.

\textbf{$\bullet$} The convergence property of our energy-constrained self-training is investigated. In addition, it can be explained from the probabilistic perspective. The throughout theoretical analysis bridge its connection with the classification expectation maximization (CEM). 

We empirically validate the effectiveness and generality of the proposed method on multiple challenging benchmarks (both classification and semantic segmentation) and achieve state-of-the-art performance.

\section{Related Works}

\noindent\textbf{Unsupervised domain adaptation (UDA)}. Deep learning (DL) has achieved tremendous milestones in computer vision recently \cite{li2019relaxed}. Instead of designing features by hand and then feeding the features to a prediction model, DL suggests learning specific features and the prediction model simultaneously from raw image following an end-to-end fashion \cite{che2019deep,liu2020severity}. However, DL is usually data-starved and relies on the $i.i.d$ assumption of training and testing data. The model trained on an annotated source domain does not generate well on a different target domain due to the domain shift (or drift) \cite{glorot2011domain,long2015learning}. In reality, collecting the large-scale labeled data in the new target domain is expensive or even prohibitive. Therefore, we would expect to migrate the knowledge from a labeled source domain to a different unlabeled target domain \cite{8578933,7410820}.

UDA with deep networks targets to learn domain invariant embeddings by minimizing the cross-domain difference of feature distributions with certain criteria. Examples of these methods include maximum mean discrepancy (MMD), deep Correlation Alignment (CORAL), sliced Wasserstein discrepancy, adversarial learning at input-level, feature level, output space level, etc. \cite{kouw2018introduction,He_2020_CVPR_Workshops}. Despite the underlying difference among these techniques, there exists an interesting connection between some of these methods with conditional forms\footnote{E.g., class-wise adversarial learning, or discriminators taking network predictions as input.} and Self-training, as they can be broadly considered as E.M. algorithms \cite{Zou_2018_ECCV}, and such conditional formulation has been widely proved to benefit the adaptation \cite{zou2019confidence}.

Other than the self-training, there are some of the other UDA methods that also rely on the pseudo label. For instance, \cite{chen2019progressive} utilizes pseudo labels to estimate target class centers, which are used to match source class centers. CAN \cite{kang2019contrastive} utilizes target pseudo-labels to estimate contrastive domain discrepancy. Since the pseudo labels can be noisy, \cite{gu2020spherical} proposes to measure its correctness by the Gaussian-uniform mixture model. Instead, we rely on our energy model for regularization. \vspace{+5pt}

\noindent\textbf{Self-training} was initially proposed for the semi-supervised learning \cite{triguero2015self}. The deep self-training is different from the self-training with fixed hand-crafted feature input in that the deep self-training involves the learning of deep embedding, which renders greater flexibility towards domain alignment than classifier-level adaptation only. With the development of deep learning and UDA, there have been several deep self-training/pseudo-label-based works that are developed for UDA \cite{busto2018open,han2019unsupervised,Zou_2018_ECCV}.

Targeting the noisy and unreliable pseudo-label, our preliminary work \cite{zou2019confidence} proposes to construct a more conservative pseudo-label that smoothes the one-hot hard label to a soft label vector. 

The EBM regularizer proposed in this work can be orthogonal to the recent progress of self-training \cite{zou2019confidence}. We resort to the additional supervision signal of EBM, which is independent of the pseudo-label. In contrast with the pre-defined label smoothing as \cite{zou2019confidence}, our proposed energy constraint is able to adaptively regularize the training with respect to the sample and the current network parameters. More appealingly, the EBM-based regularization can be simply added on state-of-the-art self-training methods following a plug-and-play fashion. \vspace{+5pt}

\noindent\textbf{Energy-Based Models (EBMs)} is developed to capture the dependencies between variables via associating scalar energy to each configuration of the variables \cite{lecun2006tutorial}. The recent studies \cite{nijkamp2019learning,grathwohl2019your} propose to approximate the expectation of the log-likelihood for a single example $\mathbf{x}$ w.r.t. the network parameter with a sampler based on the Stochastic Gradient Langevin Dynamics (SGLD) \cite{welling2011bayesian}.

EBM is a typical unsupervised generative model. To combine the classifier and EBMs, \cite{xie2016theory,du2019implicit} reinterpreted the logits (i.e., network prediction before the softmax) to define a class-conditional EBM $p(\mathbf{x}|\mathbf{y})$, which requires additional parameters to be learned to derive a classifier and an unconditional model. \cite{song2018learning} inherits a similar idea as well, while it is trained with a GAN-like generator and targets different applications. In contrast, we do not focus on the generative model of $p(\mathbf{x}|\mathbf{y})$, and propose to utilize the probability density of $p(\mathbf{x})$ as regularizer. Based on \cite{nijkamp2019learning}, the recent study \cite{grathwohl2019your} scales the training of EBMs to high-dimensional data with the Contrastive Divergence and SGLD. 

In this paper, we propose to regard EBMs as the pseudo-label-irrelevant optimization constraint, and explore their potentials on unsupervised domain adaptation (UDA). This helps realize the potential of EBMs on downstream supervised discriminative problems. In addition, thorough convergence analysis and theoretical connection with CEM are provided.

\section{Methodology}

In the scenario of UDA, we can access to the labeled source samples $(\mathbf{x}_s,\mathbf{y}_s)$ from source domain $\{\mathbf{X}_S, \mathbf{Y}_S\}$, and target samples $\mathbf{x}_t$ from unlabeled target domain data $\mathbf{X}_T$. Any target label $\hat{\mathbf{y}}_t=(\hat{y}_t^{(1)},...,\hat{y}_t^{(K)})$ from $\hat{\mathbf{Y}}_T$ is unknown. $K$ is the total number of classes. The parametric network $f_{\mathbf{w}}:\mathbb{R}^D\rightarrow \mathbb{R}^K$ with the weights $\mathbf{w}$ is used to process the $D$-dim input sample. The output $K$ real-valued numbers known as logits and usually followed by a softmax normalization to produce ${p_\mathbf{w}}(k|\mathbf{x})={\rm exp}({f_{\mathbf{w}}}({\mathbf{x}})[k])/{\sum_{k\in K}} {\rm exp} (f_{\mathbf{w}}({\mathbf{x}})[k])$ as the classifier's softmax probability for class $k$. $f_{\mathbf{w}}({\mathbf{x}})[k]$ indicates the $k^{th}$ index of $f_{\mathbf{w}}({\mathbf{x}})$, i.e., the logit corresponding the the $k^{th}$ class label.

\subsection{{Preliminary of energy-based model}}

{The energy based function $E_{\mathbf{w}}(\mathbf{x}):\mathbb{R}^D\rightarrow\mathbb{R}$ maps each point of an input space to a single scalar, which is called “energy”. One can parameterize an EBM using any function that takes $\mathbf{x}$ as the input and returns a scalar. The learning is designed in such a way that $E_{\mathbf{w}}$ can assign low energies to observed configurations of variables while give high energies to unobserved ones \cite{nijkamp2019learning}. With $E_{\mathbf{w}}$, the probability density $p(\mathbf{x})$ for $\mathbf{x}\in\mathbb{R}^D$ in the EBMs is formulated as}
\begin{align}
    p_{\mathbf{w}}(\mathbf{x}) =\frac{\exp (-E_{\mathbf{w}}(\mathbf{x}))}{Z(\mathbf{w})}\label{ebm}
\end{align} 
{where $Z(\mathbf{w})=\int_{\mathbf{x}}{\rm exp}(-E_{\mathbf{w}}(\mathbf{x}))$ is the normalizing constant (with respect to $\mathbf{x}$) also known as the partition function. Usually, the normalized densities $p_{\mathbf{w}}(\mathbf{x})$ are intractable, since we cannot reliably estimate $Z(\mathbf{w})$ for the most choice of $E_{\mathbf{w}}(\mathbf{x})$. The typical solutions rely on the sophisticate Markov Chain Monte Carlo (MCMC) sampler to train EBMs \cite{nijkamp2019learning}. In our work, we use the deep neural networks as $E_{\mathbf{w}}$. The gradient estimation is usually used for EBM optimization, and sample data from it by MCMC \cite{nijkamp2019learning}.}

\subsection{EBM for Discriminative Model}

The recent study \cite{grathwohl2019your} reveals that a standard classifier can be interpreted as an energy based model of joint distribution $p_{\mathbf{w}}(\mathbf{x},\mathbf{y})$, and optimizing the likelihood ${\rm log}p_{\mathbf{w}}(\mathbf{x},\mathbf{y})={\rm log}p_{\mathbf{w}}(\mathbf{x})+{\rm log}p_{\mathbf{w}}(\mathbf{y}|\mathbf{x})$ can be helpful for both the discrimination and generation task. Specifically, optimizing $p(\mathbf{y}|\mathbf{x})$ is simply achieved by using the standard cross-entropy loss as conventional classification. The additional optimization objective ${\rm log}p_{\mathbf{w}}(\mathbf{x})$ has been proven and demonstrating that can improve the confidence calibration and robustness for conventional classification task \cite{grathwohl2019your}.  

Considering the target samples do not have a ground truth label, the self-training methods \cite{zou2019confidence} utilize the inaccurate pseudo label to calculate the cross-entropy loss. Therefore, optimizing ${\rm log}p_{\mathbf{w}}(\mathbf{x})$ can potentially be more helpful for UDA setting. Actually, ${\rm log}p_{\mathbf{w}}(\mathbf{x})$ is adaptive w.r.t. the input $\mathbf{x}$ and network parameter $\mathbf{w}$, and irrelevant to the inaccurate pseudo label, which can be an ideal regularizer of self-training based UDA. 

However, how to modeling ${\rm log}p_{\mathbf{w}}(\mathbf{x})$ can be a challenging task. Considering $\frac{\partial{\rm log}p_{\mathbf{w}}(\mathbf{x})}{\partial{\mathbf{w}}}$ can be approximated with $-\frac{\partial E_{\mathbf{w}}(\mathbf{x})}{\partial{\mathbf{w}}}$ \cite{grathwohl2019your}, it is possible to modeling the energy function $E_{\mathbf{w}}(\mathbf{x})$ instead of ${\rm log}p_{\mathbf{w}}(\mathbf{x})$.

As detailed in Eq. (\ref{ebm}), in EBMs, the probability density $p_{\mathbf{w}}(\mathbf{x})$ for $\mathbf{x}\in\mathbb{R}^D$ can be formulated as $p_{\mathbf{w}}(\mathbf{x})=$ ${\rm exp}(-E_{\mathbf{w}}(\mathbf{x}))/Z(\mathbf{w})$. Following \cite{grathwohl2019your}, we can define an EBM of the joint distribution $p_{\mathbf{w}}(\mathbf{x},\mathbf{y})={\rm exp}(f_{\mathbf{w}}({\mathbf{x}})[k])/Z({\mathbf{w}})$, by defining $E_{\mathbf{w}}(\mathbf{x},\mathbf{y})=-f_{\mathbf{w}}({\mathbf{x}})[k]$. By marginalizing out $\mathbf{y}$, we have $p_{\mathbf{w}}(\mathbf{x})=\frac{\sum_k{\rm exp}(f_{\mathbf{w}}({\mathbf{x}})[k])}{Z({\mathbf{w}})}$ \cite{grathwohl2019your}. Considering $p_{\mathbf{w}}(\mathbf{x})={\rm exp}(-E_{\mathbf{w}}(\mathbf{x}))/Z(\mathbf{w})$, the energy function of $\mathbf{x}$ can be

\begin{align}\label{energy} E_{\mathbf{w}}(\mathbf{x})&=-{\rm LogSumExp}_k(f_{\mathbf{w}}({\mathbf{x}})[k])=-{\rm log}\sum_k{\rm exp}{(f_{\mathbf{w}}({\mathbf{x}})[k])}
\end{align}

In this setting, $p_{\mathbf{w}}(\mathbf{y}|\mathbf{x})=\frac{p_{\mathbf{w}}(\mathbf{x},\mathbf{y})}{p_{\mathbf{w}}(\mathbf{x})}=\frac{{\rm exp}(f_{\mathbf{w}}({\mathbf{x}})[k])/Z({\mathbf{w}})}{{\sum_k{\rm exp}(f_{\mathbf{w}}({\mathbf{x}})[k])}/{Z({\mathbf{w}})}}$. The normalization constant $Z({\mathbf{w}})$ will be canceled out and yielding the standard softmax function, which bridges the EBMs and conventional classifiers. Since we can factorize the likelihood as \begin{align}\label{fact}{\rm log}p_{\mathbf{w}}(\mathbf{x},\mathbf{y})={\rm log}p_{\mathbf{w}}(\mathbf{x})+{\rm log}p_{\mathbf{w}}(\mathbf{y}|\mathbf{x}) \end{align}
we can simultaneously optimize $p(\mathbf{y}|\mathbf{x})$ with standard cross-entropy loss and ${\rm log}p_{\mathbf{w}}(\mathbf{x})$ with SGLD (where gradients are taken w.r.t. ${\rm LogSumExp}_k(f_{\mathbf{w}}({\mathbf{x}})[k])$) in the supervised training setting  \cite{grathwohl2019your}.

Compared with conventional classifier which will not be affected by shifting the logits $f_{\mathbf{w}}({\mathbf{x}})$ with a scalar, the ${\log}p_{\mathbf{w}}(\mathbf{x})$ is sensitive to this kind of perturbation. It introduces an extra degree of freedom w.r.t. logits to define the density function over input examples and the joint density among examples and labels, therefore further constraining the training.

\subsection{EBM as regularization for UDA}

The self-training-based UDA proposes to utilize the iterative loss minimization scheme to gradually refine the model for the target domain. In self-training, the pseudo-labels are regarded as the learnable latent variables \cite{zou2019confidence}. Considering the target domain samples do not have the ground truth label, the optimization objective in self-training relies on the cross-entropy loss with the pseudo label, which is the relatively confident prediction in the previous round \cite{Zou_2018_ECCV}. Considering the inaccuracy of pseudo label, the smoothed soft label \cite{zou2019confidence} is proposed as an alternative to the one-hot hard pseudo label \cite{Zou_2018_ECCV}. 

Therefore, a straightforward solution for adapting EBM to UDA is to integrate Eq. \ref{energy} as a regularization term. For the target sample, the network is enforced to achieve a good prediction of pseudo-label and minimize the energy of $E_{\mathbf{w}}(\mathbf{x}_t)$. The energy constraint is inherently dependent on $\mathbf{x}$ and $\mathbf{w}$. Therefore, it is more flexible than the pre-defined label smoothing or entropy regularization \cite{zou2019confidence}. Given that the pseudo-label is usually noisy, the latter objective is expected to be even more helpful than the supervised learning task.

{Following the notations in our CRST \cite{zou2019confidence}, the self-training with EBM regularization (R-EBM) for target sample, i.e., $E_{\mathbf{w}}(\mathbf{x}_t)$, can be formulated as} \begin{align}\label{cbst}
&\underset{\mathbf{w},{{{\hat{\mathbf{Y}}}}_{T}}}{\mathop{\min }}\,\mathcal{L}_{R-EBM}(\mathbf{w}, {\hat{\mathbf{Y}}}) = -\sum\limits_{{{s}}\in {{S}}}{\sum\limits_{k=1}^{K}{y_{s}^{(k)}}\log p_\mathbf{w}(k|{\mathbf{x}_{s}})} \nonumber \\
&-\sum\limits_{{{t}}\in {{T}}}\{{\sum\limits_{k=1}^{K}{[\hat{y}_{t}^{(k)}}\log p_\mathbf{w}(k|{\mathbf{x}_{t}})}-\hat{y}_{t}^{(k)}{{\log\lambda }_{k}}]-\alpha E_{\mathbf{w}}(\mathbf{x}_t)\}  \nonumber \\
&s.t.~{{{\hat{\mathbf{y}}}}_{t}}\in \Delta^{K-1}\cup \{\mathbf{0}\},\forall t 
\end{align} For each class $k$, $\lambda_{k}$ is determined by the confidence value selecting the most confident $p$ portion of class $k$ predictions in the entire target set \cite{Zou_2018_ECCV}. If a sample's predication is relatively confident with $p_\mathbf{w}(k^*|\mathbf{x}_t) > \lambda_{k^*}$, it is selected and labeled as class $k^* = \text{argmax}_{k}\{\frac {p_\mathbf{w}(k|{\mathbf{x}_{t}})}{\lambda_k}\}$. The less confident ones with $p_\mathbf{w}(k^*|\mathbf{x}_t) \leq \lambda_{k^*}$ are not selected. The same class-balanced $\lambda_{k}$ strategy introduced in \cite{Zou_2018_ECCV} is adopted for all self-training methods in this work. 

The feasible set is the union of $\{\mathbf{0}\}$ and a probability simplex $\Delta^{K-1}$  \cite{Zou_2018_ECCV}. $\alpha$ is a balancing hyper-parameter of the regularization term, which does not directly relate to the label $\hat{y}_{t}$. The self-training can be solved by an alternating optimization scheme.\vspace{+5pt}

\noindent \textbf{Step 1) Pseudo-label generation} \label{_a)} ~ Fix $\mathbf{w}$ and solve: \begin{align}\label{cbst_a}
& \underset{{{{\hat{\mathbf{Y}}}}_{T}}}{\mathop{\min }}  -\sum\limits_{{{t}}\in {{T}}}\{{\sum\limits_{k=1}^{K}{\hat{y}_{t}^{(k)}}[\log {p_\mathbf{w}(k|{\mathbf{x}_{t}})}-\log{{{\lambda }_{k}}}]}-\alpha E_{\mathbf{w}}(\mathbf{x}_t) \} \nonumber\\
& s.t.\text{ }{{{\hat{y}}}_{t}}\in\Delta^{K-1}\cup \{\mathbf{0}\},\forall t \end{align} 

For solving step \textbf{1)}, there is a global optimizer for arbitrary $\hat{\mathbf{y}}_t=(\hat{y}_t^{(1)},...,\hat{y}_t^{(K)})$ as \cite{zou2019confidence}:
\begin{equation}
\hat{y}_{t}^{(k)*}=\left\{
\begin{aligned}
1, &~\text{if}~k={~}^{\rm argmax}_{~~~~k}~{\frac {p_\mathbf{w}(k|{\mathbf{x}_{t}})}{\lambda_k}} ~~~ \text{and} ~~~ p_\mathbf{w}(k|\mathbf{x}_t)>\lambda_k\\
0, &~\mathrm{otherwise}
\end{aligned}
\right.\label{cbst_a_solver}
\end{equation}

Despite a long period of little progression, there have been some recent works \cite{du2019implicit,nijkamp2019learning} using the sampler based on SGLD to train the large-scale EBMs on high-dimensional data, which is parameterized by deep neural networks.

Given that the goal of this work is to incorporate EBM training into the standard classification setting, the classification part is the same as \cite{zou2019confidence}, and we only change the regularizer. Therefore, we can follow \cite{grathwohl2019your} to train the network with both cross-entropy loss and SGLD to ensure this distribution is being optimized with an unbiased objective. Similar to \cite{du2019implicit} we also adopt the contrastive divergence to estimate the expectation of derivative of the log-likelihood for a single example $\mathbf{x}$ with respect to $\mathbf{w}$. Since it gives an order of magnitude savings in computation compared to seeding new chains at each iteration as in \cite{nijkamp2019learning}.\vspace{+5pt}

\noindent \textbf{Step 2) Network retraining} \label{b)} ~ Fix $\hat{\mathbf{Y}}_T$ and minimize \begin{align}\tiny\label{cbst_b}
-\sum\limits_{{{s}}\in {{S}}}{\sum\limits_{k=1}^{K}{y_{s}^{(k)}}\log p_\mathbf{w}(k|{\mathbf{x}_{s}})}-\sum\limits_{{{t}}\in {{T}}}{\sum\limits_{k=1}^{K}{\hat{y}_{t}^{(k)}}\log p_\mathbf{w}(k|{\mathbf{x}_{t}})} 
\end{align}

\noindent w.r.t. $\mathbf{w}$. Carrying out step \textbf{1)} and \textbf{2)} for one time is defined as one round in self-training.

For each class $k$, $\lambda_{k}$ is determined by the confidence value selecting the most confident $p$ portion of class $k$ predictions in the entire target set \cite{Zou_2018_ECCV}. If a sample's predication is relatively confident with $p(k^*|\mathbf{x}_t;\mathbf{w}) > \lambda_{k^*}$, it is selected and labeled as class $k^{*}=\argmax_{k}\{\frac {p(k|{\mathbf{x}_{t}};\mathbf{w})}{\lambda_k}\}$. The less confident ones with $p(k^*|\mathbf{x}_t;\mathbf{w}) \leq \lambda_{k^*}$ are not selected. The same class-balanced $\lambda_{k}$ strategy introduced in \cite{Zou_2018_ECCV} is adopted for all self-training methods in this work.

\subsection{EBM loss for target domain}

The essence of the energy-based model \cite{lecun2006tutorial} is a data-driven process that shapes the energy surface in such a way that the desired configurations get assigned low energies, while the incorrect ones are given high energies.

In this framework, the conventional supervised learning \cite{lecun2005loss} for each $\mathbf{x}$ in the training set, the energy of the pair $\{\mathbf{x,y}\}$ takes low values when $\mathbf{y}$ is the correct label and higher values for incorrect $\mathbf{y}$. Similarly, when modeling $\mathbf{x}$ alone within
an unsupervised learning setting, lower energy is attributed to the data manifold. Actually, the term contrastive sample is often used to refer to a data point causing an energy pull-up, such as the incorrect $\mathbf{y}$ in supervised learning and points from low data density regions in the setting of unsupervised learning. {Therefore, we can construct an efficient EBM loss for the target sample as} \begin{equation}\label{eq6}\mathcal{L}_{EBM}=-{\rm log}\sum_{{k}} {\hat{\mathbf{y}}} {\rm exp}(f_{\mathbf{w}}({\mathbf{x}})[{{k}}])+{\sum_{{k}}{\hat{y}_{t}^{(k)}}\log{{{\lambda }_{k}}}}
\end{equation}

The soft pseudo-label ${\hat{\mathbf{y}}}$ used here is the original softmax prediction of the previous round classifier/segmentor without the one-hot processing. This is also essentially different from the soft-label used in \cite{zou2019confidence} which post-process the one-hot label to construct a smoothed one. Then, Eq. \ref{eq6} will be used to replace the objective in Eq. \ref{cbst_a}. This configuration also dispenses with the validation of $\alpha$.

This can be regarded as using ${\hat{\mathbf{y}}}$ as the weight to indicate which class's energy should be minimized. For the class with larger ${\hat{\mathbf{y}}}$ probability, the larger $f_{\mathbf{w}}({\mathbf{x}})[{{k}}]$ will be emphasised. Therefore, the network will pron to produce the large $f_{\mathbf{w}}({\mathbf{x}})[{{k}}]$ value at the pseudo-label's category. 

With the soft pseudo-label setting, the confidence of each class prediction can also be encoded in the histogram distribution of ${\hat{\mathbf{y}}}$. Although the additional information can enrich the training, the signal may be even noisier at the beginning. To avoid the model collapse, in practice, we propose a loss annealing scheme that starts with the EBM regularization and then smoothly changes to $\mathcal{L}_{EBM}$. The objective can be formulated as ${\frac{1}{1+\beta}\left\{\mathcal{L}_{EBM}+\beta \mathcal{R}_{EBM}\right\}}$ where $\beta$ is a balance hyper-parameter and decrease $\beta$ from 10 to 0 gradually in the training, $i.e.,$ $\beta=\frac{10}{1+N^2}$ for $N\leq$5, and $\beta=0$ for $N>$5, where $N$ is the epoch number.

\subsection{Energy-based self-training as CEM}

\cite{amini2002semi} proves that semi-supervised logistic regression is a Classification Expectation-Maximization (CEM) to solve Classification Maximum Likelihood (CML). We generalize the statement for proposed energy-constrained self-training as follows.

\begin{proposition} Energy-constrained self-training can be a regularized Classification  Expectation-Maximization (CEM) for solving a generalized Regularized Classification Maximum Likelihood (RCML) problem and is convergent with gradient descent.\end{proposition}

The problem of classification maximum likelihood (CML) was first proposed for clustering tasks in \cite{celeux1992classification}, and can be solved through classification expectation maximization (CEM). Compared with traditional expectation maximization (EM) that has an Expectation (E) step and a Maximization (M) step, CEM has an additional Classification (C) step (between E and M step) that assigns a sample to the cluster with maximal posterior probability. In \cite{amini2002semi}, the CML is generalized for discriminant semi-supervised learning with both labeled and unlabeled data that is defined as follows:
\begin{equation}\label{cml}
{\log{\mathcal{L}}_{C}} = {\log\tilde{\mathcal{L}}_{C}}  +\sum\limits_{{{i}}\in {{S,T}}}\log p(\mathbf{x}_i) 
\end{equation}
where:
\begin{equation}\label{ecml}
\begin{split}
{\log\tilde{\mathcal{L}}_{C}} = \sum\limits_{{{s}}\in {{S}}}{\sum\limits_{k=1}^{K}{y_{s}^{(k)}}\log p_\mathbf{w}(k|{\mathbf{x}_{s}})}+\sum\limits_{{{t}}\in {{T}}}{\sum\limits_{k=1}^{K}{\hat{y}_{t}^{(k)}}\log p_\mathbf{w}(k|{\mathbf{x}_{t}})} 
\end{split}
\end{equation}
Note that ${{{\hat{y}}}_{t}}\in\{0,1\}^K, \forall t$. $p_\mathbf{w}(k|\mathbf{x}_t)$ is a posterior probability modeled by probabilistic classifier such as logistic classifier, neural network, etc. $\mathbf{w}$ is the learnable weight. As mentioned in \cite{amini2002semi}, when using a discriminant classifier, we make no assumption about the data distribution $p(\mathbf{x}_t)$. Thus maximizing Eq. \ref{cml} is equal to maximizing Eq. \ref{ecml}. The Classification Expectation Maximization (CEM) is used to solve Eq. \ref{ecml} via alternating E-step: estimating the posterior probability $p_\mathbf{w}(k|\mathbf{x}_t)$; C-step: assigning the pseudo-labels according to the maximal posterior probability $p_\mathbf{w}(k|\mathbf{x}_t)$; M-step: maximizing the log-likelihood in terms of $\mathbf{w}$.

{For proposition 1, we show the inherent connections between the energy-based self-training and Regularized Classification Maximum Likelihood (RCML). We also show that the proposed alternating optimization is essentially a Classification Expectation-Maximization (CEM) towards maximizing RCML. The energy-based self-training minimization problem can be seen as the following maximization problem}:
\begin{equation}\label{cspst}
\begin{split}
& \underset{{{{\mathbf{w},\hat{\mathbf{Y}}}}_{T}}}{\mathop{\max }}\,{\log\mathcal{\tilde{L}}_{C}}+EBM =\sum\limits_{{{s}}\in {{S}}}{\sum\limits_{k=1}^{K}{y_{s}^{(k)}}\log(p_\mathbf{w}(k|{\mathbf{x}_{s}}))}\\
&~~~~~~~~~ +\sum\limits_{{{t}}\in {{T}}}[{\sum\limits_{k=1}^{K}{\hat{y}_{t}^{(k)}}\log (p_\mathbf{w}(k|{\mathbf{x}_{t}}))}-\alpha E_{\mathbf{w}}(\mathbf{x}_t)]\\ 
&\text{s.t.}~~\mathbf{\hat{y}}_t\in~\Delta^{(K-1)}\cup \{\mathbf{0}\},~\forall t
\end{split}
\end{equation}
{where $\alpha>0$. $\mathbf{\hat{y}}_t\in\Delta^{K-1}=\{(\hat{y}_{t}^{(1)},..,\hat{y}_{t}^{(K)}):\hat{y}_{t}^{(k)}\ge 0,\sum\limits_{k=1}^{K}{\hat{y}_{t}^{(k)}}=1\}, \forall t$. ($\Delta^{K-1}$ denotes a probability simplex) Compared with Eq. \ref{ecml}, the above problem can be regarded as a generalized Classification Maximization Likelihood ($\log\tilde{\mathcal{L}}_{C}$) with a energy regularizer ($E$), relaxing the one-hot vector feasible set to the probability simplex. The alternating optimization can be regarded as a Classification Expectation Maximization (CEM) for solving this regularized CML (RCML) problem:}

\noindent\textbf{E-Step:} Given the model weight $\mathbf{w}$, estimate the posterior probability $p_\mathbf{w}(\mathbf{x}_t),\forall t$.

\noindent\textbf{C-Step:} Fix $\mathbf{w}$ and solve the following problem for $\hat{\mathbf{Y}}_T$.
\begin{equation}\label{rcml_a}
\begin{split}
&\underset{{{{\hat{\mathbf{Y}}}}_{T}}}{\mathop{\max }}\,\sum\limits_{{{t}}\in {{T}}}[{\sum\limits_{k=1}^{K}{\hat{y}_{t}^{(k)}}\log p_\mathbf{w}(k|{\mathbf{x}_{t}})}-\alpha E_{\mathbf{w}}(\mathbf{x}_t)]\\ 
&\text{s.t.}~~\hat{\mathbf{y}}_t\in~\Delta^{(K-1)}\cup \{\mathbf{0}\},~\forall t
\end{split}
\end{equation} 

\noindent\textbf{M-Step:} Fix $\hat{\mathbf{Y}}_T$ and use gradient ascent to solve the following problem for $\mathbf{w}$.
\begin{equation}\label{rcml_b}
\begin{split}
\underset{{{{\mathbf{w}}}}}{\mathop{\max }}\,&\sum\limits_{{{s}}\in {{S}}}{\sum\limits_{k=1}^{K}{y_{s}^{(k)}}\log p_\mathbf{w}(k|{\mathbf{x}_{s}})}+\sum\limits_{{{t}}\in {{T}}}[{\sum\limits_{k=1}^{K}{\hat{y}_{t}^{(k)}}\log p_\mathbf{w}(k|{\mathbf{x}_{t}})}-\alpha E_{\mathbf{w}}(\mathbf{x}_t)]
\end{split}
\end{equation}
Now we have shown energy-based self-training can be regarded as a CEM.

{In the next subsection, we also provide the convergence analysis of energy-constrained self-training for UDA.}

\subsection{Convergence analysis for energy-constrained self-training}\label{crst_converge}

{Considering that our energy-constrained self-training follows an alternating optimization scheme, which is different from the conventional single round updating, we further explain that the energy-constrained self-training is convergent. We emphasize the proposed energy regularizer is convex w.r.t. either prediction distributions or pseudo-label distributions. The problem can be solved by alternatively taking the two steps. The training is convergent only if the loss in both of the two steps are non-increasing.} 

Specifically, we have the following analysis for these two steps:

\noindent\textbf{Loss in step (1) is non-increasing:} We prove that the solver (\ref{cbst_a_solver}) is a global minimum of optimization problem (\ref{cbst_a}). Specifically, the feasible set of the problem is a union of $\Delta^{K-1}$ and $\{\textbf{0}\}$. Thus, two sub-problems share the same objective with either a probability simplex and $\textbf{0}$ as a feasible set. For the proposed convex energy regularizer, the first sub-problem is a convex problem with a unique solution (see Table 1 of the main paper), while the latter one has a global optimum solver at $\textbf{0}$. Therefore, (\ref{cbst_a_solver}) is the solver for the combined two subproblems, making it the global optimum. Thus step 1 is non-increasing and can be minimized with the global solver.

\noindent\textbf{Loss in step (2) is non-increasing:} Due to the convexity of proposed model regularizers, with a proper learning rate, the gradient descent can monotonically decrease the regularized objective function of (\ref{cbst_b}).

Since the proposed energy regularization are lower-bounded by $0$, it's easy to see that energy regularized self-training loss is lower-bounded by $\sum\limits_{t\in T}\sum\limits_{{{k=1},...,{K}}}\log\lambda_k$. Thus energy-regularized self-training is convergent.

\begin{table*}[t]
\centering
\resizebox{\linewidth}{!}{
\centering
\begin{tabular}{c|cccccccccccc|c}
ine
Method         &  Aero & Bike   & Bus & Car & Horse & Knife & Motor   & Person   & Plant & Skateboard & Train  & Truck   & Mean \\ ineine
Source-Res101 \cite{gholami2019taskdiscriminative} & 55.1 & 53.3 & 61.9 & 59.1 & 80.6 & 17.9 & 79.7 & 31.2 & 81.0 & 26.5 & 73.5 & 8.5 & 52.4 \\ine

MMD \cite{long2015learning} & 87.1 & 63.0 & 76.5 & 42.0 & 90.3 & 42.9 & 85.9 & 53.1 & 49.7 & 36.3 & 85.8 & 20.7 & 61.1 \\
DANN \cite{ganin2016domain} & 81.9 & 77.7 & 82.8 & 44.3 & 81.2 & 29.5 & 65.1 & 28.6 & 51.9 & 54.6 & 82.8 & 7.8 & 57.4 \\ 
ENT \cite{grandvalet2005semi} & 80.3 & 75.5 & 75.8 & 48.3 & 77.9 & 27.3 & 69.7 & 40.2 & 46.5 & 46.6 & 79.3 & 16.0 & 57.0 \\
MCD \cite{saito2017maximum} & 87.0 & 60.9 & \textbf{83.7} & 64.0 & 88.9 & 79.6 & 84.7 & {76.9} & {88.6} & 40.3 & 83.0 & 25.8 & 71.9 \\
ADR \cite{saito2018adversarial} & 87.8 & 79.5 & \textbf{83.7} & 65.3 & \textbf{92.3} & 61.8 & {88.9} & 73.2 & 87.8 & 60.0 & {85.5} & {32.3} & 74.8 \\  
DEV \cite{you2019toward} & 81.8& 53.5& 82.9& 71.6& 89.2 &72.0& {89.4}& 75.7& \textbf{97.0} & 55.5 &71.2 &29.2& 72.4\\
TDDA \cite{gholami2019taskdiscriminative} & 88.2 &78.5& 79.7& 71.1 &90.0 &81.6& 84.9 &72.3 &92.0& 52.6& 82.9& 18.4 &74.03\\

PANDA  \cite{hu2020panda} &90.9 &50.5 &72.3& \textbf{82.7} &88.3 &\textbf{88.3} &\textbf{90.3}& 79.8& 89.7 &79.2& \textbf{88.1} &39.4& 78.3\\

DMRL \cite{wu2020dual} & - &-& -&- &- &-& - &- &-& -& -& - &75.5\\ine

CBST \cite{Zou_2018_ECCV} & 87.1$\pm$1.2 & 79.5$\pm$2.3 & 58.3$\pm$2.6 & 50.4$\pm$3.9 & 82.8$\pm$2.1 & 73.7$\pm$7.2 & 80.9$\pm$2.6 & 71.8$\pm$3.1 & 81.6$\pm$3.2 & 88.4$\pm$3.3 & 75.2$\pm$1.2 & 68.4$\pm$3.4 & 74.8$\pm$0.5 \\

ine

\textbf{CBST+$R_{EBM}$} & 87.9$\pm$1.6 & 79.6$\pm$1.5 & 68.5$\pm$1.3 & 68.6$\pm$1.9 & 83.2$\pm$1.2 & {78.4$\pm$1.9} & 83.5$\pm$1.5 & 72.2$\pm$1.5 & 82.2$\pm$1.6 & 84.3$\pm$1.5 & 80.9$\pm$1.4 & 67.5$\pm$1.3 & 77.0$\pm$0.6 \\

\textbf{CBST+${\mathcal{L}}_{EBM}$} & 87.8$\pm$1.2 & 80.4$\pm$1.9 & 69.3$\pm$1.7 & 67.5$\pm$1.6 & 83.2$\pm$1.0 & 76.8$\pm$1.4 & 81.6$\pm$1.8 & 72.3$\pm$1.5 & 81.6$\pm$1.8 & 84.5$\pm$1.3 & 81.7$\pm$1.1 & 67.0$\pm$1.4 & 76.8$\pm$0.5  \\

ine

CRST\cite{zou2019confidence} & 89.2$\pm$1.6 & {79.6$\pm$4.6} & 64.2$\pm$4.0 & 57.8$\pm$3.4 & 87.8$\pm$1.9 & 79.6$\pm$8.5 & 85.6$\pm$2.6 & 75.9$\pm$4.2 & 86.5$\pm$2.2 & 85.1$\pm$2.4 & 77.7$\pm$2.2 & 68.5$\pm$0.9 & {78.1$\pm$0.7} \\

ine

\textbf{CRST+$R_{EBM}$} & 90.3$\pm$1.5 & \textbf{82.6$\pm$1.2} & 72.4$\pm$1.5 & {71.7$\pm$1.8} & 87.6$\pm$1.8& {80.5$\pm$1.9} & 85.4$\pm$1.5 & \textbf{80.8$\pm$1.5} & 87.1$\pm$1.6 & \textbf{89.9$\pm$1.5} & 83.6$\pm$1.6 & \textbf{71.5$\pm$1.3 }& {80.2$\pm$0.5}\\

\textbf{CRST+${\mathcal{L}}_{EBM}$} & \textbf{91.9$\pm$1.7} & 81.8$\pm$1.8 & 72.8$\pm$1.2 & 71.5$\pm$1.3 & 88.4$\pm$1.6 & {81.3$\pm$1.3} & {89.3$\pm$1.1} & 78.8$\pm$1.2 & 87.2$\pm$1.7 & 88.0$\pm$1.5 & 85.3$\pm$1.0 & 71.2$\pm$1.6 & \textbf{80.4$\pm$0.7} \\

ine
ine

SimNet* \cite{pinheiro2018unsupervised} & \textbf{94.3} & 82.3 & 73.5 & 47.2 & 87.9 & 49.2 & 75.1 & 79.7 & 85.3 & 68.5 & 81.1 & 50.3 & 72.9 \\

GTA* \cite{sankaranarayanan2018generate}  
& - & - & - & - & - & - & - & - & - & - & - & - & 77.1\\ine

\textbf{CRST+$R_{EBM}$* }& {93.2$\pm$1.3} & \textbf{85.8$\pm$1.2} &  {73.7$\pm$1.0 }& \textbf{74.3$\pm$1.5 }& {89.5$\pm$0.6} & \textbf{87.6$\pm$1.6} &  {88.2$\pm$1.6 }& \textbf{82.2$\pm$1.6 }& \textbf{90.9$\pm$1.3} & \textbf{91.6$\pm$1.8} & {85.1$\pm$1.4} & \textbf{79.9$\pm$1.4} & \textbf{82.8$\pm$0.5 }\\

\textbf{CRST+${\mathcal{L}}_{EBM}$*} & {93.0$\pm$1.6} & {84.4$\pm$1.6} & \textbf{74.4$\pm$1.0} & {73.5$\pm$1.4} & \textbf{91.2$\pm$1.5} & {82.4$\pm$1.5} & \textbf{88.8$\pm$1.6} & {81.8$\pm$1.2 }& 89.8$\pm$1.2 & 90.8$\pm$1.4 & \textbf{86.5$\pm$1.2} &  {77.1$\pm$0.9 }& {82.5$\pm$0.5}\\

ine

\end{tabular}%
}
\caption{{Experimental results for VisDA17-val setting. We use ResNet101 as backbone except SimNet and GTA.*ResNet152 backbone.}}
\label{table:visda17}
\end{table*}


\begin{figure*}[t]
\centering
\includegraphics[width=12cm]{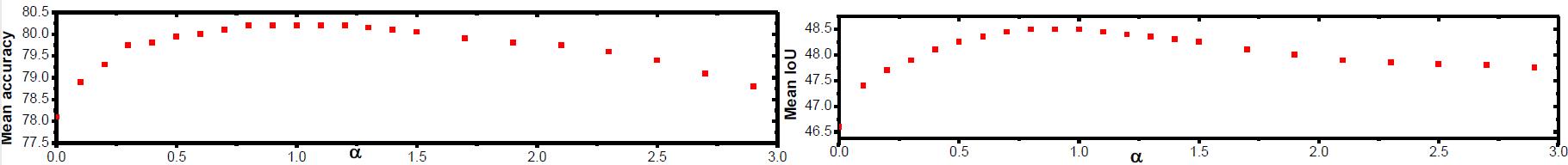}\\
\caption{Sensitive analysis of hyper-parameter $\alpha$ in VisDA17 (top) and CTA52Sityscapes (bottom) with CRST+$R_{EBM}$.}\label{fig:3} \vspace{+10pt}
\end{figure*}

\begin{table*}[t]
\centering
\resizebox{1\linewidth}{!}{%
\begin{tabular}{c|cccccc|c}
ine
Method & A$\rightarrow$W & D$\rightarrow$W & W$\rightarrow$D & A$\rightarrow$D & D$\rightarrow$A & W$\rightarrow$A & Mean Accuracy \\ ine
ResNet-50 \cite{he2016deep} & 68.4$\pm$0.2 & 96.7$\pm$0.1 & 99.3$\pm$0.1 & 68.9$\pm$0.2 & 62.5$\pm$0.3 & 60.7$\pm$0.3 & 76.1 \\ineine

DAN [ICML2015] \cite{long2015learning} & 80.5$\pm$0.4 & 97.1$\pm$0.2 & 99.6$\pm$0.1 & 78.6$\pm$0.2 & 63.6$\pm$0.3 & 62.8$\pm$0.2 & 80.4 \\
RTN [NeurIPS2016] \cite{long2016unsupervised} & 84.5$\pm$0.2 & 96.8$\pm$0.1 & 99.4$\pm$0.1 & 77.5$\pm$0.3 & 66.2$\pm$0.2 & 64.8$\pm$0.3 & 81.6 \\
DANN [JMLR2016] \cite{ganin2016domain} & 82.0$\pm$0.4 & 96.9$\pm$0.2 & 99.1$\pm$0.1 & 79.7$\pm$0.4 & 68.2$\pm$0.4 & 67.4$\pm$0.5 & 82.2 \\
ADDA [CVPR2017] \cite{tzeng2017adversarial} & 86.2$\pm$0.5 & 96.2$\pm$0.3 & 98.4$\pm$0.3 & 77.8$\pm$0.3 & 69.5$\pm$0.4 & 68.9$\pm$0.5 & 82.9 \\
JAN [JMLR2017] \cite{long2017deep} & 85.4$\pm$0.3 & 97.4$\pm$0.2 & 99.8$\pm$0.2 & 84.7$\pm$0.3 & 68.6$\pm$0.3 & 70.0$\pm$0.4 & 84.3 \\
GTA [CVPR2018] \cite{sankaranarayanan2018generate} & {89.5$\pm$0.5} & 97.9$\pm$0.3 & 99.8$\pm$0.4 & 87.7$\pm$0.5 & 72.8$\pm$0.3 & 71.4$\pm$0.4 & 86.5 \\ 

DMRL [ECCV2020] \cite{wu2020dual}  &\textbf{90.8$\pm$0.3} & 99.0$\pm$0.2  &\textbf{100.0$\pm$0.0} & \textbf{93.4$\pm$0.5} & 73.0$\pm$0.3 & 71.2$\pm$0.3  &87.9\\

ineine

CBST [ECCV2018] \cite{Zou_2018_ECCV}& 87.8$\pm$0.8 & 98.5$\pm$0.1 & \textbf{100$\pm$0.0} & 86.5$\pm$1.0 & 71.2$\pm$0.4 & 70.9$\pm$0.7 & 85.8 \\ine

CBST+$R_{EBM}$ & 89.4$\pm$0.2 & 99.6$\pm$0.1 & \textbf{100$\pm$0.0} & 88.7$\pm$0.9 & 73.8$\pm$0.2 &  {73.1$\pm$0.2} & 87.4\\

CBST+${\mathcal{L}}_{EBM}$  & 89.0$\pm$0.4 & 99.6$\pm$0.1 & \textbf{100$\pm$0.0} & 88.4$\pm$0.8 & \textbf{74.7$\pm$0.2} & 73.0$\pm$0.4 & 87.4 \\ineine

CRST [ICCV2019] \cite{zou2019confidence}& 89.4$\pm$0.7 &  {98.9$\pm$0.4} & \textbf{100$\pm$0.0} & 88.7$\pm$0.8 & 72.6$\pm$0.7 & 70.9$\pm$0.5 & {86.8} \\ine

CRST+$R_{EBM}$  &  {90.6$\pm$0.4} & \textbf{99.7$\pm$0.1} & \textbf{100$\pm$0.0} & {91.0$\pm$0.8} & {74.0$\pm$0.6} & \textbf{74.0$\pm$0.3} & \textbf{88.6} \\

CRST+${\mathcal{L}}_{EBM}$ & 90.4$\pm$0.9 & \textbf{99.7$\pm$0.1} & \textbf{100$\pm$0.0} & 90.0$\pm$0.9 & 73.7$\pm$0.8 & 73.9$\pm$0.4 & 88.3  \\ine

\end{tabular}%
}\vspace{+3pt}
\caption{{Experimental results for Office-31}}
\label{tabel:office}
\end{table*}

\section{Experiments}

This section provides comprehensive evaluations of our proposed EBM-based regularization and loss on several UDA benchmarks, including image classification and semantic segmentation. We implement our methods with the PyTorch platform \cite{paszke2017automatic} on a V100GPU.

\subsection{Domain Adaptation for Image Classification}

We test on two challenging benchmarks: VisDA17 \cite{peng2018visda} and Office-31 \cite{saenko2010adapting}. 

The VisDA17 \cite{peng2018visda} dataset involves 12-class for UDA classification problem. Following the standard protocol as \cite{zou2019confidence,sankaranarayanan2018generate}, the source domain utilizes the training set with $152,409$ synthetic 2D images, while the target domain uses the validation set with $55,400$ real images from the COCO dataset.

For a fair comparison, we use the same backbone network for VisDA17 as the compared works, e.g., ResNet-101 and ResNet-152. Moreover, the model was also pre-trained with the ImageNet dataset as the previous works and fine-tuned in source domain by SGD \cite{zou2019confidence}. We use a fixed learning rate $1\times10^{-3}$, weight decay $5\times10^{-4}$, momentum $0.9$ and batch size $32$. From Fig. \ref{fig:3}, we can see that the performance is not sensitive to $\alpha$ when $\alpha\in[0.8,1.1]$. Therefore, we simply choose $\alpha=1$ consistently for all settings.

\begin{table*}[t]
\centering
\resizebox{\linewidth}{!}{
\centering
\begin{tabular}{c|c|ccccccccccccccccccc|c}
ine
Method         & Base Net          & Road & SW   & Build & Wall & Fence & Pole & TL   & TS   & Veg. & Terrain & Sky  & PR   & Rider & Car  & Truck & Bus  & Train & Motor & Bike & mIoU \\ ineine
Source     & DRN26 & 42.7 & 26.3 & 51.7  & 5.5  & 6.8   & 13.8 & 23.6 & 6.9  & 75.5 & 11.5    & 36.8 & 49.3 & 0.9   & 46.7 & 3.4   & 5.0  & 0.0   & 5.0   & 1.4  & 21.7 \\
CyCADA \cite{hoffman2018cycada}         &   & 79.1 & 33.1 & 77.9  & 23.4 & 17.3  & 32.1 & 33.3 & 31.8 & 81.5 & 26.7 & 69.0 & 62.8 & 14.7  & 74.5 & 20.9  & 25.6 & 6.9   & 18.8  & 20.4 & 39.5 \\ ine
Source     & DRN105          &  36.4 & 14.2 & 67.4 & 16.4 & 12.0 & 20.1 & 8.7 & 0.7 & 69.8 & 13.3 & 56.9 & 37.0 & 0.4 & 53.6 & 10.6 & 3.2 & 0.2 & 0.9 & 0.0 & 22.2 \\
MCD \cite{saito2017maximum}   &   &90.3 & 31.0 & 78.5 & 19.7 & 17.3 & 28.6 & 30.9 & 16.1 & 83.7 & 30.0 & 69.1 & 58.5 & 19.6 & 81.5 & 23.8 & 30.0 & 5.7 & 25.7 & 14.3 & 39.7 \\ ine
Source & PSPNet & 69.9 & 22.3 & 75.6 & 15.8 & 20.1 & 18.8 & 28.2 & 17.1 & 75.6 & 8.00 & 73.5 & 55.0 & 2.9 & 66.9 & \textbf{34.4} & 30.8 & 0.0 & 18.4 & 0.0 & 33.3 \\
DCAN \cite{Wu_2018_ECCV} & & 85.0 & 30.8 & 81.3 & 25.8 & 21.2 & 22.2 & 25.4 & 26.6 & 83.4 & {36.7} & 76.2 & 58.9 & 24.9 & 80.7 & 29.5 & 42.9 & 2.50 & 26.9 & 11.6 & 41.7 \\ ine

Source     & DeepLabv2          &  75.8 & 16.8 & 77.2 & 12.5 & 21.0 & 25.5 & 30.1 & 20.1 & 81.3 & 24.6 & 70.3 & 53.8 & 26.4 & 49.9 & 17.2 & 25.9 & 6.5 & 25.3 & 36.0 & 36.6\\
AdaptSegNet \cite{Tsai_adaptseg_2018}   & & 86.5 & 36.0 & 79.9& 23.4 & {23.3} & 23.9 & 35.2 & 14.8 & 83.4 & 33.3 & 75.6 & 58.5 & 27.6 & 73.7 & 32.5 & 35.4 & 3.9 & 30.1 & 28.1 & 42.4 \\ ine

AdvEnt \cite{vu2019advent}   & DeepLabv2 & 89.4 & 33.1 & 81.0 & 26.6 & 26.8 & 27.2 & 33.5 & 24.7 & 83.9 & 36.7 & 78.8 & 58.7 & 30.5 & \textbf{84.8} & 38.5 & 44.5 & 1.7 & 31.6 & 32.4 & 45.5 \\ ine
Source     & DeepLabv2          &  - & - & - & - & - & - & - & - & -& - & - & - & - & - & - & - & - & - & - & 29.2\\
FCAN \cite{zhang2018fully}   &  &  - & - & - & - & - & - & - & - & -& - & - & - & - & - & - & - & - & - & - &  {46.6} \\ ine

Source&DeepLabv2 &75.8 &16.8 &77.2 &12.5 &21.0&25.5 &30.1 &20.1 &81.3 &24.6 &70.3& 53.8& 26.4 &49.9 &17.2 &25.9& 6.5& 25.3 &36.0 &36.6\\

DPR \cite{tsai2019domain} & &92.3 &51.9 &82.1 &29.2& 25.1& 24.5& 33.8& \textbf{33.0} &82.4& 32.8& \textbf{82.2}& 58.6& 27.2& 84.3 &33.4& \textbf{46.3} &2.2 &29.5 &32.3& 46.5\\ine

Source&DeepLabv2 &73.8& 16.0& 66.3& 12.8 &22.3& 29.0& 30.3 &10.2 &77.7 &19.0 &50.8 &55.2& 20.4& 73.6& 28.3 &25.6 &0.1& 27.5 &12.1 &34.2\\

PyCDA \cite{lian2019constructing} && 90.5&  36.3&  84.4&  32.4 & \textbf{28.7} & 34.6&  36.4 & 31.5&  86.8&  37.9&  78.5 & 62.3 & 21.5&  85.6&  27.9 & 34.8&  18.0 & 22.9&  \textbf{49.3} & 47.4\\ine
ine

Source &DeepLabv2 & 71.3 & 19.2 & 69.1 & 18.4 & 10.0 & 35.7 & 27.3 &  6.8 & 79.6 & 24.8 & 72.1 & 57.6 & 19.5 & 55.5 & 15.5 & 15.1 & 11.7 & 21.1 & 12.0 & 33.8 \\

CBST \cite{Zou_2018_ECCV}      &                   & 89.9 &  55.0 &  79.9 &  29.5 &  20.6 &  37.8 &  32.9 &  13.9 &  84.0 &  31.2 &  75.5 &  60.2 &  27.1 &  81.8 &  29.7 &  40.5 &  7.62 &  28.7 &  41.4 & 45.6 \\

CBST+$R_{EBM}$      &   & 91.1 & 53.9 & 80.6 & 31.6 & 21.0 & \textbf{40.4} & 35.0 & 19.8 & \textbf{86.6} & 35.9 & 76.4 &{63.3}& \textbf{31.4} & 83.0 & 22.5 & 38.6 & 24.2 & 32.2 & 39.4 & 47.8 \\

CBST+${\mathcal{L}}_{EBM}$    &                   & \textbf{92.7} & 56.2 & 79.6 & 27.1 & 21.7 & 39.3 & 35.5 & 24.2 & 85.3 & \textbf{38.0} & 77.9 & 63.6 & 29.1 & 84.8 & 25.3 & 39.6 & 17.8 & \textbf{34.8} & 43.4 & 48.2 \\ ineine

Source &DeepLabv2 & 71.3 & 19.2 & 69.1 & 18.4 & 10.0 & 35.7 & 27.3 &  6.8 & 79.6 & 24.8 & 72.1 & 57.6 & 19.5 & 55.5 & 15.5 & 15.1 & 11.7 & 21.1 & 12.0 & 33.8 \\

CRST \cite{zou2019confidence}    &                   & 89.0 & 51.2 & 79.4 & 31.7 & 19.1 & 38.5 & 34.1 & 20.4 & 84.7 & 35.4 & {76.8} & 61.3 & {30.2} & 80.7 & 27.4 & 39.4 & 10.2 & 32.2 & {43.3} & {46.6} \\

CRST+$R_{EBM}$    &         & 92.5 & \textbf{56.6} & 80.9 & 26.2 & 20.5 & 40.2 & 35.3 & 24.4 & \textbf{86.6} &  {37.3} & 77.5 & 63.4 & {30.5} & 81.3 & 28.8 & 39.2 & 20.6 & {33.5} & 41.3 & 48.5\\

CRST+${\mathcal{L}}_{EBM}$   &         & \textbf{92.7} & 55.1 & \textbf{83.6} & \textbf{34.5} & 20.4 & {39.4} & \textbf{36.5} & 19.6 &  {85.7} & 35.7 & 78.3 & \textbf{64.2}
& 30.1 &  {84.5} & 32.0 & 40.4 & \textbf{28.0} &  {33.9} & 44.2 & \textbf{49.0}  \\ ine

\end{tabular}%
}
\caption{{Experimental results for GTA5 to Cityscapes.}}
\label{table:gtacity}
\end{table*}

We present the results on VisDA 17 in Table \ref{table:visda17} regarding per-class accuracy and mean accuracy. For each proposed approach, we independently run five times, and report the average accuracy and the corresponding standard deviation ($\pm$sd).

Among the compared methods, Class-balanced self-training (CBST) and Conservative regularized self-training (CRST) \cite{zou2019confidence} are the state-of-the-art self-training UDA methods. Our regularization can be simply added on CBST/CRST. We denote the CRST with EBM regularization with CRST+$R_{EBM}$.

Contributed by the additional EBMs objective, our proposed CBST+$R_{EBM}$ and CRST+$R_{EBM}$  can outperform CBST and CRST significantly. More appealingly, it can outperform the recent UDA methods other than self-training by a large margin. More appealing, the adversarial training can be simply added on to further boost the performance of self-training \cite{zou2019confidence}. Our proposed energy-based constraint is able to adaptively regularize the training w.r.t. the input $\mathbf{x}$ and the present network parameters, which is more flexible than the manually pre-defined label smoothing \cite{zou2019confidence}.

We can see that CRST+$R_{EBM}$ achieves a better performance than the recent adversarial training \cite{gholami2019taskdiscriminative}, dropout \cite{saito2017adversarial} and moment matching methods, which revokes the potential of self-training in UDA.

To demonstrate the generality of our EBM-based regularizer and loss for different backbones, we also tested with the more powerful backbones that have also been applied and shown better results \cite{pinheiro2018unsupervised,sankaranarayanan2018generate}. The EBM objective with the ResNet152 backbone outperforms the other state-of-the-arts \cite{pinheiro2018unsupervised,sankaranarayanan2018generate}. It also demonstrates the flexibility of $R_{EBM}$ for different backbones.


Office-31 is another standard UDA dataset containing images belonging to $31$ classes from three different domains, i.e., Amazon (A), Webcam (W), and DSLR (D), each containing $2,817$, $795$ and $498$ images respectively. We follow the standard protocol in \cite{saenko2010adapting,sankaranarayanan2018generate} and evaluate on six transfer tasks $A \rightarrow W$, $D \rightarrow W$, $W \rightarrow D$, $A \rightarrow D$, $D \rightarrow A$, and $W \rightarrow A$.

\begin{figure}[t]
\centering
\includegraphics[width=12cm]{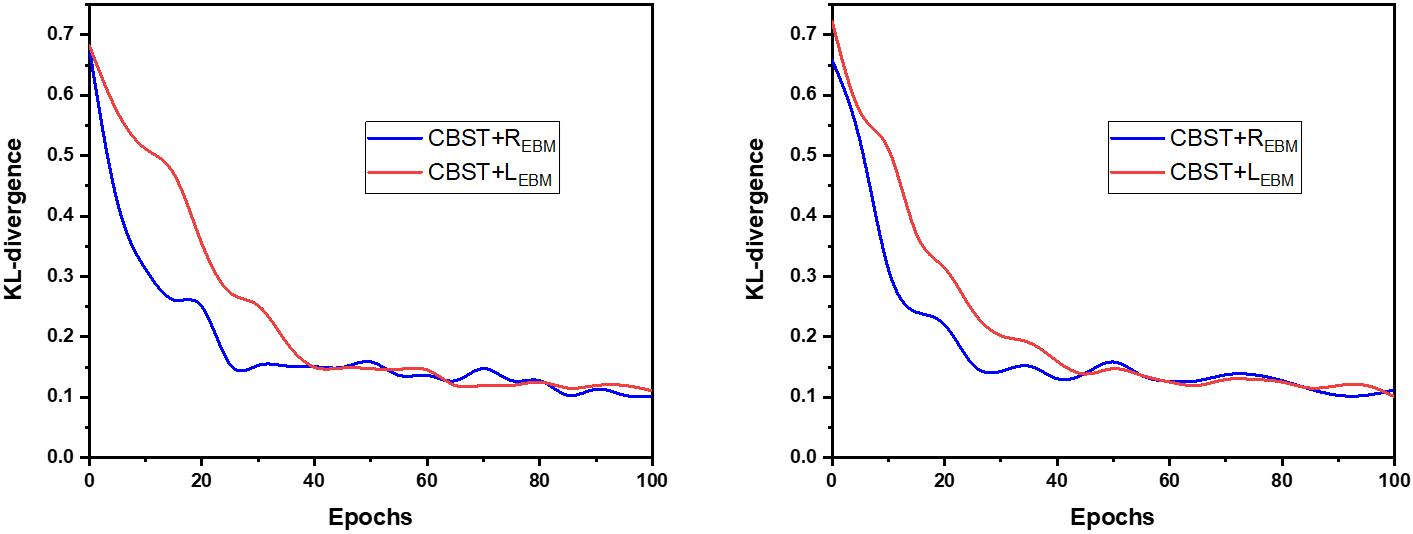}\\
\caption{The KL-divergence between the predicted $p_t(y)$ and the ground of truth $p_t(y)$ in the testing stage. Left: VisDA17 task with CBST backbone, right: GTA5 to Cityscapes task with CRST backbone. The small value indicates the more accurate estimation.}\label{fig:4} 
\end{figure}

We compare the performances of different methods on Office-31 with the same basenet ResNet-50 in Table 4. CBST or CRST+$R_{EBM}$ achieves the best performance that outperforms their baselines CBST and CRST.

Actually, our $R_{EBM}$ can be orthogonal with the recent advanced self-training UDA methods. CBST \cite{Zou_2018_ECCV} is a pioneer of vanilla adversarial UDA, and its performance on VisDA17 is reported on \cite{zou2019confidence}. We note that the label smoothing or entropy regularization proposed in CRST \cite{zou2019confidence} or the vanilla self-training UDA \cite{Zou_2018_ECCV} can be simply add-on our $R_{EBM}$ to further improve the performance. The proposed CRST+ $R_{EBM}$ obtains better or competitive performances in all of the settings. Especially, it is able to outperform the sophisticated generative pixel-level domain adaptation method GTA in both architecture and objectives.

\subsection{UDA for Semantic Segmentation}

\begin{table*}[t]
\centering
\resizebox{\linewidth}{!}{%
\begin{tabular}{c|c|cccccccccccccccc|c|c}
ine
Method      & Base Net          & Road & SW   & Build & Wall* & Fence* & Pole* & TL   & TS   & Veg. & Sky  & PR   & Rider & Car  & Bus  & Motor & Bike & mIoU & mIoU* \\ ine
Source     & DRN105          &  14.9 & 11.4 & 58.7 & 1.9 & 0.0 & 24.1 & 1.2 & 6.0 & 68.8 & 76.0 & 54.3 & 7.1 & 34.2 & 15.0 & 0.8 & 0.0 & 23.4 & 26.8 \\
MCD \cite{saito2017maximum}   &   & \textbf{84.8} & \textbf{43.6} & 79.0 & 3.9 & 0.2 & 29.1 & 7.2 & 5.5 & 83.8 & 83.1 & 51.0 & 11.7 & 79.9 & 27.2 & 6.2 & 0.0 & 37.3  & 43.5 \\ ine
Source & PSPNet & 56.0 & 24.6 & 76.5 & 5.0 & 0.2 & 19.0 & 5.7 & 7.8 & 77.5 & 78.9 & 44.7 & 7.70 & 35.3 & 7.9 & 1.5 & 24.0 & 29.5 & 34.5 \\
DCAN \cite{wu2018dcan} & & 82.8 & 36.4 & 75.7 & 5.1 & 0.1 & 25.8 & 8.0 & 18.7 & 74.7 & 76.9 & 51.1 & 15.9 & 77.7 & 24.8 & 4.11 & 37.3 & 38.4 & 44.9 \\ ine
Source only & DeepLabv2       &  55.6 & 23.8 & 74.6 & $-$ & $-$ & $-$ & 6.1 & 12.1 & 74.8 & 79.0 & 55.3 & 19.1 & 39.6 & 23.3 & 13.7 & 25.0 & $-$ & 38.6\\
AdaptSegNet \cite{Tsai_adaptseg_2018} &    & 84.3 & 42.7 & 77.5 & $-$ & $-$ & $-$ & 4.7 & 7.0 & 77.9 & \textbf{82.5} & 54.3 & 21.0 & 72.3 & 32.2 & 18.9 & 32.3 & $-$ & 46.7 \\ ine
Source only & ResNet-38         & 32.6 & 21.5 & 46.5 & 4.8 & 0.1 & 26.5 & 14.8 & 13.1 & 70.8 & 60.3 & 56.6 & 3.5   & 74.1 & 20.4 & 8.9  & 13.1 & 29.2 & 33.6\\
CBST \cite{Zou_2018_ECCV}        &                   & 53.6 & 23.7 & 75.0 & 12.5 & 0.3 & 36.4 & {23.5} & 26.3 &  {84.8} & 74.7 & \textbf{67.2} & 17.5 &  {84.5} & 28.4 & 15.2 & \textbf{55.8} & 42.5 & 48.4\\

CBST+$R_{EBM}$        &                   & 77.8 & 37.8 & 75.1 & 17.0 & \textbf{10.8} & 36.3 & 20.9 & 32.8 & 78.2 & 79.4 & 55.7 & 21.9 & 82.2 & 13.0 & 20.6 & 45.3 & 43.6 & 49.6 \\
CBST+${\mathcal{L}}_{EBM}$      &                   & 65.7 & 26.4 & 78.1 & \textbf{19.0} & 3.2 & 31.5 & 21.9 & 32.5 & 81.4 & 78.4 & 56.8 & \textbf{29.8} & 82.3 & 29.7 &  {21.0} & 45.4 & 42.8 & 49.8 \\

ine

Source only & DeepLabv2         & 64.3 & 21.3 & 73.1 & 2.4 & 1.1 & 31.4 & 7.0 & 27.7 & 63.1 & 67.6 & 42.2 & 19.9 & 73.1 & 15.3 & 10.5 & 38.9 & 34.9 & 40.3 \\
CBST    \cite{Zou_2018_ECCV}       &                   & 68.2 & 32.0 & 76.6 & 15.1 &  {2.5} & 29.2 & 20.7 &  {33.8} & 73.2 & 78.1 & 60.0 & 25.0 & 83.5 & 15.1 & 20.2 & 40.2 & 42.1 & 48.8 \\
CRST \cite{zou2019confidence}       &                   & 65.4 & 27.7 & \textbf{80.1} &  {18.3} & 0.7 & {38.1} & 22.4 & 30.8 & 83.2 & 80.0 & 60.6 & 27.5 & 82.6 &  {33.5} & 19.0 & 42.8 &  {44.6} & {50.4} \\

CRST+$R_{EBM}$       &                   & 67.6 & 24.9 & 78.9 & 18.0 & 9.2 & 36.9 & \textbf{26.9} & \textbf{35.5} & 82.4 & 79.4 & 59.5 & \textbf{29.8} & \textbf{85.0} & 29.9 & \textbf{22.0} & 47.4 & \textbf{45.7} & 51.7 \\

CRST+${\mathcal{L}}_{EBM}$    &                   & 64.6 & 30.3 & 78.7 & 16.8 & 5.8 & \textbf{39.0} & 25.2 & 31.8 & \textbf{85.6} & 80.7 & 64.5 &  {29.3} & 83.5 & \textbf{35.3} & 20.1 & 44.5 & 45.6 & \textbf{51.8} \\ 

ine
\end{tabular}%
}
\caption{{Experimental results of SYNTHIA $\rightarrow$ Cityscapes}}\label{table:syncity}
\end{table*}


The typical deep semantic segmentation is essentially making the pixel-wise classification. We consider the challenging segmentation adaptation settings in the synthetic-to-real scenario: 

1) GTA5 \cite{richter2016playing} to Cityscapes \cite{cordts2016cityscapes}. There are 19 shared classes between these two datasets. GTA5 dataset has 24,966 annotated images with the size of 1,052$\times$1,914, which are rendered by the GTA5 game engine. 

2) SYNTHIA \cite{ros2016synthia} to Cityscapes \cite{cordts2016cityscapes}. There are 16 shared classes between these two datasets. We use the SYNTHIA-RAND-CITYSCAPES set, which consists of $9,400$ labeled images. Following the standard protocols in \cite{hoffman2018cycada,Tsai_adaptseg_2018}, we use the full set of SYNTHIA/GTA5, and propose to adapt the segmentation model to the Cityscapes training set with $2,975$ images. In testing, we evaluate on the Cityscapes validation set with $500$ images. 

To make a fair comparison with the other methods, we use the ResNet101 as the backbone network as \cite{Zou_2018_ECCV,zou2019confidence}. Noticing that in PSPNET \cite{zhao2017pyramid}, Wide ResNet38 is a stronger basenet than ResNet101. The basenet is pre-trained in ImageNet and fine-tuned in source domain by SGD with learning rate $2.5\times10^{-4}$, weight decay $5\times10^{-4}$, momentum $0.9$, batch size $2$, patch size $512\times1024$ and data augmentation of multi-scale training ($0.5 - 1.5$) and horizontal flipping. All results on this dataset in the main paper are unified to report the mIoU of the models at the end of the $6^{th}$ epoch.

We compare CBST/CRST+$R_{EBM}$ with the other methods in Table \ref{table:gtacity}. Based on the previous self-training UDA methods CBST or CRST, our additional EBM objective outperforms CBST or CRST by about 2\% w.r.t. the mean IoUs. 

We achieve a new state-of-the-art, even compared with the generative pixel-level domain adaptation method GTA \cite{zhang2018fully}, which is a relatively complex algorithm in both architecture and objectives. As shown in Fig. \ref{fig:4}, $R_{EBM}$ converges stably along the iterations. The performance on SYNTHIA to Cityscapes is also consistent with the GTA5 to Cityscapes as shown in Table \ref{table:syncity}.


\section{Conclusions}

In this paper, we introduced the EBM’s objective into the self-training UDA. Considering the lack of the target label, we resort to the pseudo-label, which is usually noisy. The EBM optimization objective provides an additional signal that is independent of the pseudo-labels. This approach can be more promising in the UDA setting than in the supervised learning setting as it can be added as either a regularization term or combined with a confidence-aware soft pseudo-label in an EBM loss. In addition, our solution is orthogonal with the recent advances in self-training, and can be added in a plug-and-play manner without large computation and change in network structure. Extensive experiments on both UDA classification and semantic segmentation demonstrated the effectiveness and generality of our approach. More advanced EBM training and self-training methods such as \cite{han2019unsupervised} can also be adapted to further improve the performance, which is subject to our future work.

\section{Data Availability Statement}
This study does not involves new data collection. All of the datasets used in this study are public available.

\section*{References}

\bibliography{mybibfile}

\end{document}